\documentclass[11pt]{article}

\usepackage[preprint]{acl}

\usepackage{times}
\usepackage{latexsym}

\usepackage[T1]{fontenc}

\usepackage[utf8]{inputenc}

\usepackage{microtype}

\usepackage{inconsolata}

\usepackage{graphicx}
\usepackage{amsmath}
\usepackage{amssymb}
\usepackage{bm}
\usepackage{booktabs}
\usepackage{makecell}
\title{UNIQUE: Universal Top-k Sparse Attention for Training-free Inference and Sparsity-aware Training}

\author{
  Keqi Deng, Shaoshi Ling, Ruchao Fan, Jinyu Li\\
  Microsoft, USA \\
  \texttt{keqideng@microsoft.com} \\
}

\begin{document}
\maketitle
\begin{abstract}

Long-context inference in large language models (LLMs) is bottlenecked by the linear growth of the self-attention key-value (KV) cache. Top-$k$ sparse attention alleviates this by loading only a small fraction of the KV cache, but accurately and cheaply estimating cache importance, for both training-free use and sparsity-aware training, remains challenging. 
This paper proposes UNIQUE, a universal top-$k$ sparse attention framework that addresses both requirements and stays consistently effective across LLM modalities. UNIQUE operates at the granularity of KV pages and estimates per-page importance with a simple yet accurate score combining the mean of the page's keys as a representative vector with their standard deviation as an offset term. To further close the train-inference gap, this paper introduces a soft-mask sparsity-aware training scheme that uses the top-$k$ score boundary as a per-query threshold and a sigmoid soft mask around it, requiring neither auxiliary losses nor architectural changes. Experiments on text and speech LLMs show that UNIQUE preserves task performance on long-context benchmarks such as LongBench Pro and on long-form speech recognition, while delivering up to 11.4$\times$ attention-kernel speedup over FlashInfer dense attention and at least 5.3$\times$ end-to-end decoding speedup over a vLLM-based dense model.
\end{abstract}

\section{Introduction}

Large language models (LLMs) have rapidly expanded their context windows from a few thousand to hundreds of thousands of tokens~\cite{achiam2023gpt,dubey2024llama,comanici2025gemini,abouelenin2025phi}, unlocking applications such as long-document understanding, repository-level code reasoning, and long-horizon multi-turn agents. This growing context length, however, makes the key value (KV) cache rather than model weights the dominant source of memory traffic during auto-regressive decoding: each generated token requires reading the entire KV cache from high-bandwidth memory, and the size of this cache grows linearly with sequence length \cite{DBLP:conf/mlsys/PopeDCDBHXAD23, deng2026multi}.
At hundred-thousand-token scales, the per-token KV cache read can dominate decoding latency and cap end-to-end throughput, regardless of how fast the matrix multiplications themselves are~\citep{dao2023flashattention2,DBLP:conf/sosp/KwonLZ0ZY0ZS23,ye2025flashinfer}. 
The same pressure is no longer confined to text. Speech and audio LLMs increasingly operate on hours-long 
recordings for meeting transcription, podcast understanding, and streaming dialogue~\citep{DBLP:conf/icml/ParkSJKRS25,peng2025vibevoice}, where a single recording can already produce a speech sequence of tens of thousands of positions and stress the KV cache in the same way as long-context text. Reducing the cost of long-context attention is therefore a shared bottleneck across modalities rather than a text-only concern.

To address the KV cache bottleneck, sparse attention drops less informative tokens to compute attention over a small context subset. Existing approaches typically follow one of two paradigms. Training-free methods~\citep{DBLP:conf/nips/Zhang00CZC0TRBW23,DBLP:conf/nips/XiaoZ0XLZ0024,DBLP:conf/icml/TangZZXKH24} dynamically select top-$k$ KV pairs without updating model weights. While they offer easy deployment, a mismatch between dense training and sparse inference often leads to notable performance degradation under high sparsity. Alternatively, trainable sparse mechanisms such as Native Sparse Attention (NSA)~\citep{DBLP:conf/acl/YuanGD0ZZXWW0WR25} and its variants~\citep{zhao2025infllm} mitigate this drop by incorporating sparsity directly into model training. 
However, these methods typically rely on specialized architectural components or auxiliary routing modules, limiting their flexibility and often demanding resource-intensive training. In practice, a unified framework that provides immediate training-free acceleration while remaining optimizable through sparsity-aware fine-tuning is more desirable. Moreover, current sparse attention strategies are almost exclusively validated on text LLMs, leaving their robustness on continuous modalities such as long-form speech in speech LLMs largely unexplored.


This paper proposes UNIQUE, a universal top-$k$ sparse attention framework that supports both training-free deployment and sparsity-aware training across modalities. For hardware efficiency, UNIQUE operates at KV-page granularity~\cite{DBLP:conf/sosp/KwonLZ0ZY0ZS23}. For each page, UNIQUE estimates its importance with a simple yet accurate score. It first computes the mean of all keys within the page as a representative vector. Considering that the mean alone may dilute the salience of a few highly informative keys, UNIQUE introduces an additional offset term based on the standard deviation of the keys, which compensates for the underestimated importance. This offset-augmented score surpasses existing training-free sparse attention methods across modalities. To further close the gap between dense training and sparse inference, the top-$k$ decision boundary serves as a per-query threshold, and a sigmoid function is applied to derive a differentiable soft mask. This allows the model to adapt to sparse attention and actively refine its top-$k$ selection during training, requiring no auxiliary loss or architectural change. Experiments on long-context text benchmarks and long-form speech recognition show that UNIQUE preserves task quality while significantly accelerating decoding.

The main contributions of this paper can be summarized in four parts:
\begin{itemize}
    \item A universal top-$k$ sparse attention framework, UNIQUE, that unifies training-free and trainable sparse attention in a single design, validated on both text and speech modalities.
    \item An offset-augmented page scoring method that combines the mean and standard deviation of keys within each page, achieving more accurate importance estimation than existing approaches.
    \item A sigmoid soft-mask scheme that derives a differentiable gate from the top-$k$ decision boundary, enabling sparsity-aware training without auxiliary losses or extra parameters.
    \item Efficient CUDA implementations including fused kernels and optimized operators, e.g., a top-$k$ selection that is at least $2\times$ faster than PyTorch and FlashInfer counterparts.
\end{itemize}

\section{Related Work}

\subsection{KV Cache Reduction}

Reducing the memory footprint of the KV cache has been extensively studied at the architecture level. MQA~\cite{shazeer2019fast} and GQA~\cite{DBLP:conf/emnlp/AinslieLJZLS23} reduce the number of key-value heads, with GQA retaining a moderate number of groups to trade off between cache size and representational capacity. MLA~\cite{liu2024deepseek} compresses all key-value heads into a single low-rank latent, effectively recovering MQA-level cache cost while surpassing GQA in expressiveness.


Orthogonal to architectural changes, token compression methods reduce KV cache size by permanently discarding or condensing tokens deemed less important.
H2O~\cite{DBLP:conf/nips/Zhang00CZC0TRBW23} evicts KV entries with low accumulated attention scores, and SnapKV~\cite{DBLP:conf/nips/LiHYVLYCLC24} prunes tokens based on observed attention patterns.
Rather than outright removal, beacon-based approaches~\cite{DBLP:conf/iclr/Zhang0XSYD25} insert summary tokens that compress preceding chunks into compact representations.
However, as analyzed in Quest~\cite{DBLP:conf/icml/TangZZXKH24}, such compression is inherently irreversible: once a token is discarded or merged, the original information becomes inaccessible to subsequent queries.
In contrast, sparse attention retains the full KV cache and dynamically selects a query-specific subset, preserving access to any past token.


\subsection{Sparse Attention}

\paragraph{Training-free methods}

\begin{figure*}[ht]
  \centering
  \includegraphics[width=\linewidth]{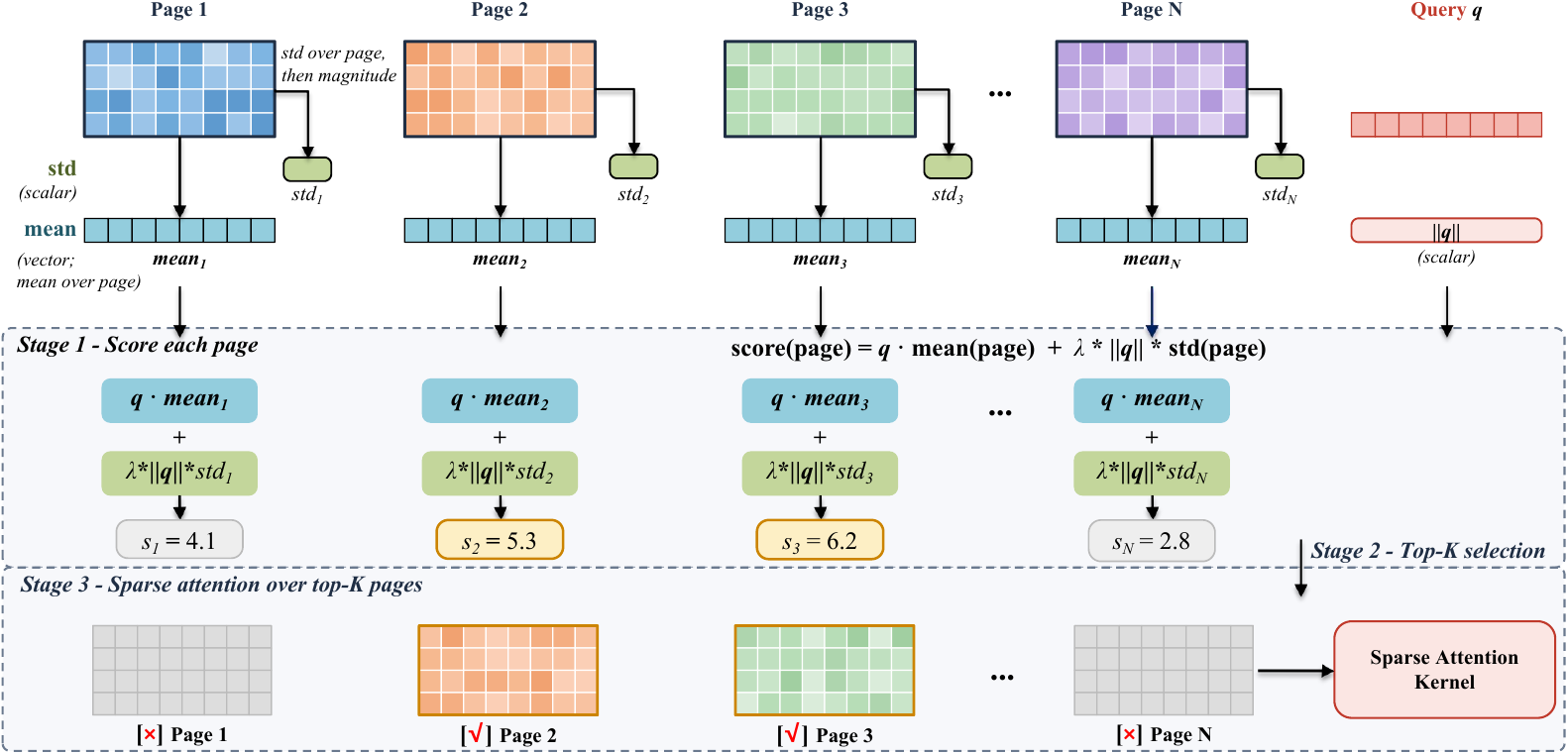}
  \vspace{-0.55cm}
  \caption{Overview of page criticality estimation in UNIQUE. For each KV page, UNIQUE precomputes a key mean (a vector) and a scalar standard deviation (std) of its keys. Stage~1 combines them with the query into a per-page score, Stage~2
selects the top-$k$ pages, and Stage~3 sends only these pages to
the sparse attention kernel.}
  \label{fig:criticality}
\end{figure*}

Training-free sparse attention exploits the inherent sparsity of attention in dense-trained models.
Quest~\cite{DBLP:conf/icml/TangZZXKH24} organizes keys into pages and selects top-$k$ pages per query via a min-max method, while InfLLM~\cite{DBLP:conf/nips/XiaoZ0XLZ0024} extends this to million-length contexts by offloading distant pages to CPU.
MInference~\cite{jiang2024minference} and FlexPrefill~\cite{DBLP:conf/iclr/LaiLLMZ25} target the prefilling phase; this paper instead focuses on decoding, which dominates end-to-end latency as reported by ~\citet{DBLP:conf/icml/TangZZXKH24} (e.g., over 86\% for 16k-token prompts with 512-token responses).
SpargeAttention~\cite{DBLP:conf/icml/ZhangXHWX0C25} and
BLASST~\cite{yuan2025blasst} skip value loading based on key
importance; BLASST is capped at $2\times$ since keys are still
fully loaded, and SpargeAttention requires per-task offline
calibration of its sparse setup.
Overall, training-free methods often stay at conservative sparsity to
avoid degradation, limiting the achievable acceleration.

\paragraph{Trainable methods}

Trainable sparse attention incorporates sparsity directly into training, allowing models to actively adapt to sparse patterns. NSA~\cite{DBLP:conf/acl/YuanGD0ZZXWW0WR25} introduces a tri-branch design with compressed, selected, and sliding attention, achieving strong performance but requiring training from scratch. InfLLM-v2~\cite{zhao2025infllm} simplifies NSA by removing the compressed branch, enabling fine-tuning from dense checkpoints.
SeerAttention~\cite{gao2025seerattention} instead trains a lightweight router via self-distillation to predict block-level sparsity.
In contrast, UNIQUE proposed in this paper requires no auxiliary loss or extra architectural components, while supporting both training-free deployment and sparsity-aware fine-tuning in a single unified design.

\section{Methodology}

This section presents UNIQUE, a page-level top-$k$ sparse attention framework that performs sparse decoding without any training, and can be further refined when sparsity-aware training data is available. The page criticality estimator at the core of UNIQUE is built from two cheap statistics of each page's keys (Section~\ref{sec:criticality}), supported by a fused pipeline that turns the abstraction into real wall-clock speedup at inference (Section~\ref{sec:inference}). When training is allowed, a differentiable sigmoid mask is applied around the top-$k$ boundary, letting the model adapt to sparsity without any auxiliary loss or extra parameters (Section~\ref{sec:training}).

\subsection{Page Criticality Estimation}
\label{sec:criticality}
UNIQUE partitions the KV cache into contiguous pages following \citet{DBLP:conf/sosp/KwonLZ0ZY0ZS23} and assigns each page a single criticality score. Let the cache contain $N$ keys at the current decoding step, partitioned into $P = \lceil N / S \rceil$ contiguous pages of size $S$. The $p$-th page is denoted as $\textbf{K}_p = \{\bm{k}_{p,1}, \ldots, \bm{k}_{p,S}\}$ with $\bm{k}_{p,j} \in \mathbb{R}^{D}$, and the current decoding query is denoted as $\bm{q} \in \mathbb{R}^{D}$. The goal of UNIQUE is to assign each page $p$ a single scalar score $\mathrm{score}(p)$, so that only the top-$k$ scoring pages are attended to during the actual attention computation. The full pipeline is illustrated in Fig.~\ref{fig:criticality}.

For every page, UNIQUE precomputes two cheap statistics that depend only on its keys and can therefore be cached once and reused across all subsequent decoding steps. The first is the per-dimension key mean of the page,
\begin{equation}
\label{eq:mean}
\bm{mean}_p \;=\; \tfrac{1}{S}\sum_{j=1}^{S} \bm{k}_{p,j} \;\in\; \mathbb{R}^{D},
\end{equation}
which serves as a single representative vector of the keys inside the page. The second statistic measures how dispersed the keys are around this mean,
\begin{equation}
\label{eq:std}
{std}_p = \Bigl\| \sqrt{\tfrac{1}{S}\sum_{j=1}^{S} (\bm{k}_{p,j} - \bm{mean}_p)^{2}} \,\Bigr\|_2 \in \mathbb{R},
\end{equation}
where the per-dimension standard deviation is first computed, and then collapsed into a single scalar per page through its $\ell_2$ norm.

Given the query $\bm{q}$, each page is then scored by combining the two precomputed statistics with a single hyperparameter $\lambda \ge 0$:
\begin{equation}
\label{eq:score}
\mathrm{score}(p) \;=\; \bm{q} \cdot \bm{mean}_p \;+\; \lambda \, \|\bm{q}\|_2 \, {std}_p,
\end{equation}
with $\lambda$ fixed to $0.5$ in all experiments. The first term $\bm{q} \cdot \bm{mean}_p$ measures how well the query aligns with the average key of the page, but the mean alone is prone to dilution: when a page contains a few highly informative keys mixed with many uninformative ones, averaging washes the informative ones out and the page receives only a moderate score. The second term $\lambda \, \|\bm{q}\|_2 \, std_p$ counteracts this by lifting the scores of pages whose keys are spread out, since such pages are more likely to hide a sharp, query-relevant key. The multiplication by $\|\bm{q}\|_2$ keeps the two terms on a comparable scale across layers and queries, without introducing any 
learnable parameter.



\subsection{Inference and Efficient Implementation}
\label{sec:inference}

At decoding time, let $\mathbf{K},\mathbf{V} \in \mathbb{R}^{N \times D}$ denote the keys and values currently stored in the KV cache and $\bm{q} \in \mathbb{R}^{D}$ denote the decoding query. The dense attention output is
\begin{equation}
\label{eq:dense_attn}
\bm{o} \;=\; \mathrm{softmax}\!\left( \frac{\bm{q} \mathbf{K}^{\!\top}}{\sqrt{D}} \right) \mathbf{V}.
\end{equation}
UNIQUE replaces this with a sparse counterpart that only reads the keys and values of the top-$k$ pages chosen by Eq.~\ref{eq:score}. Let $\mathcal{T}_k(\bm{q})$ be the indices of those pages, and let $\mathbf{K}_{\mathcal{T}_k(\bm{q})},\mathbf{V}_{\mathcal{T}_k(\bm{q})} \in \mathbb{R}^{kS \times D}$ stack their keys and values along the sequence axis. The sparse attention output is then
\begin{equation}
\label{eq:sparse_attn}
\bm{o} \;=\; \mathrm{softmax}\!\left( \frac{\bm{q} \mathbf{K}_{\mathcal{T}_k(\bm{q})}^{\!\top}}{\sqrt{D}} \right) \mathbf{V}_{\mathcal{T}_k(\bm{q})}.
\end{equation}
Each decoding step of UNIQUE thus decomposes into three CUDA operations: criticality estimation, top-$k$ selection, and sparse attention. 

\paragraph{Fused criticality estimation kernel}
Since GQA is adopted in this paper, pages are partitioned along the KV-head dimension, and the top-$k$ selection is performed independently for each of the KV heads $H_{kv}$. 
Within each group, the $G = H / H_{kv}$ query heads $\{\bm{q}_g\}_{g \in \mathcal{G}}$ sharing the same KV head produce $G$ raw scores against every page, and UNIQUE aggregates them by taking the maximum across the group:
\begin{equation}
\label{eq:score_gqa}
\mathrm{score}(p) \;=\; \max_{g \in \mathcal{G}}\!\left[\, \bm{q}_g \cdot \bm{mean}_p \;+\; \lambda \, \|\bm{q}_g\|_2 \, \mathrm{std}_p \,\right],
\end{equation}
so a page is kept if any query head in the group finds it relevant, consistent with GQA semantics.

A naive realization of Eq.~\ref{eq:score_gqa} performs three separate launches: a batched matmul for the $G \times P$ raw scores per KV head, an elementwise add for the per-head offset $\lambda \, \|\bm{q}_g\|_2 \, \mathrm{std}_p = \lambda \, (\|\bm{q}\|_{1:G}) (\mathrm{std}_{1:P})^{\top}$, which is a rank-one outer product, and a max reduction across the $G$ query heads. Each launch round-trips the $G \times P$ intermediate score tensor through high-bandwidth memory (HBM), which dominates the cost when $P$ is large. UNIQUE fuses all three steps into a single CUDA kernel: a batched matmul produces the $G \times P$ raw scores, a fused epilogue applies the rank-one offset and the max reduction without ever materializing the full $G \times P$ offset tensor in HBM, and only the final $P$-vector of per-page scores is written back. This roughly halves the kernel's memory traffic.
On an H100 GPU, this fusion alone yields up to \textbf{1.73$\times$} speedup over the naive unfused implementation. More details refer to Appendix~\ref{app:fused_kernel_bench}.

\paragraph{Radix-based top-$k$ selection}

Once the per-page scores are available, UNIQUE extracts the $k$ largest of them for each KV head. Generic top-$k$ routines in deep-learning frameworks leave room for
improvement here when $k \ll P$ and the subsequent sparse attention
does not need the top-$k$ pages in sorted order. UNIQUE therefore
reformulates the problem as a two-round 8-bit radix selection over
bf16 scores: each round builds a $256$-bin histogram in shared
memory, locates the bucket containing the $k$-th largest element,
and shrinks the search range by one byte. With only two rounds the
total work is $O(P)$ instead of $O(P\log P)$, and the entire
computation for one row fits in a single CUDA block. The
implementation is specialized for three regimes: for $P \le 4096$
all scores are cached in registers and shared memory is reserved
for the histograms; for larger $P$ the scores are staged in shared
memory, which accommodates up to $P \approx 23$K pages under the
default $48$\,KB per-block budget shared by all recent NVIDIA GPUs; beyond that, UNIQUE falls back to the top-$k$
routine of FlashInfer~\citep{ye2025flashinfer}. Selected physical
page IDs are gathered via the page table inside the same
kernel, so no intermediate index tensor reaches HBM. On an H100
GPU this implementation runs up to $2.0\times$ faster
than \texttt{flashinfer.top\_k} and $4.8\times$ faster than \texttt{torch.topk}; see Appendix~\ref{app:radix_topk_bench} for details.

\paragraph{Sparse attention kernel}
After top-$k$ selection, the sparse attention reduces to a standard paged attention call: for each query head, the keys and values of the selected pages are read from the KV cache and attended to. Since UNIQUE selects whole pages of contiguous keys and values, each selected page can be loaded with coalesced memory accesses by the attention kernel, exactly as in standard paged attention~\citep{ye2025flashinfer}, 
where the selected pages are streamed from HBM into shared memory and $\bm{q}\mathbf{K}^{\top}$, softmax, and reweighting of $\mathbf{V}$ are performed entirely on chip. The page-level granularity of UNIQUE composes naturally with this kernel: switching from dense to sparse paged attention requires only passing in a different set of page IDs, and the cost of this step scales with the token budget $k \cdot S$ rather than the KV length $N$. 


\subsection{Sparsity-aware Training}
\label{sec:training}

While UNIQUE already works well in the training-free setting, it further supports a sparsity-aware training scheme that not only adapts the model to sparse attention but also trains the criticality estimation itself, without any auxiliary loss or architectural change. For brevity, let $s_p \equiv \mathrm{score}(p)$, and let $s_{(1)} \ge s_{(2)} \ge \cdots \ge s_{(P)}$ denote these scores sorted in descending order. The midpoint of the $k$-th and $(k{+}1)$-th largest scores is taken as the per-query top-$k$ boundary,
\begin{equation}
\label{eq:boundary}
\theta \;=\; \mathrm{sg}\!\left(\frac{s_{(k)} + s_{(k+1)}}{2}\right),
\end{equation}
where $\mathrm{sg}(\cdot)$ stops the gradient.
A differentiable soft mask is defined for each page as
\begin{equation}
\label{eq:soft_mask}
g_p \;=\; \sigma\!\left(\frac{\mathrm{score}(p) - \theta}{\tau}\right),
\end{equation}
where the query norm $\|\bm{q}\|_2$ inside $\mathrm{score}(p)$ is detached from gradient for training stability, and $g_p$ is applied as an additive log-bias $\log g_p$ inside the attention softmax at training, so that the gradient flows back to the page scores through $g_p$.

This boundary choice concentrates training signal where it is most useful. The derivative $\partial g_p / \partial s_p = g_p(1 - g_p)/\tau$ peaks at $s_p = \theta$ and decays rapidly away from it, so pages that are clearly inside or outside the top-$k$ receive negligible gradient while pages near the boundary are pushed apart. Sparsity-aware training therefore widens the margin around the top-$k$ cutoff rather than perturbing already-saturated scores. 
At the end of training, the soft mask can be switched to a hard top-$k$ mask for a small number of steps, which further aligns training with the exact selection used at inference.

The page-level soft mask $g_p \in \mathbb{R}^{Q \times P}$ is much
smaller than the $\mathbb{R}^{Q \times N}$ mask used in token-level
schemes, and is fused into a FlashAttention-style kernel that
performs the gate, the softmax, and their backward passes in a
single block-wise pass without materializing the full attention
matrix. Sparsity-aware fine-tuning therefore runs at wall-clock
cost comparable to dense fine-tuning while still propagating
gradients to the page scores.


\section{Experiments}
\label{sec:exp}

\begin{table*}[t]
\centering
\small
\renewcommand\arraystretch{1.03}
\setlength{\tabcolsep}{9pt}
\begin{tabular}{l|c|c|cc|cccc}
\Xhline{3\arrayrulewidth}
\textbf{Method} & \textbf{Budget} & \textbf{Overall} & \textbf{En} & \textbf{Zh} & \textbf{Extreme} & \textbf{Hard} & \textbf{Moderate} & \textbf{Easy} \\
\hline
Full attention & --   & 37.70 & 35.88 & 38.96 & 31.13 & 33.72 & 31.60 & 48.86 \\
\hline
Quest          & 2048 & 30.63 & 29.17 & 32.60 & 24.60 & 28.60 & 24.17 & 42.01 \\
H2O            & 2048 & 33.02 & 31.77 & 34.42 & 27.51 & 31.08 & 26.56 & 43.33 \\
InfLLM         & 2048 & 36.16 & 35.06 & 37.53 & 29.95 & 32.90 & 29.84 & \textbf{47.98} \\
Proposed \textbf{UNIQUE}& 2048 & \textbf{37.11} & \textbf{35.57} & \textbf{38.15} & \textbf{31.28} & \textbf{33.18} & \textbf{31.03} & 47.66 \\
\hline
Quest          & 1024 & 27.18 & 25.83 & 28.76 & 21.53 & 22.69 & 21.62 & 38.71 \\
H2O            & 1024 & 31.61 & 30.59 & 33.16 & 26.93 & 29.15 & 25.43 & 41.87 \\
InfLLM         & 1024 & 35.42 & 34.77 & 36.68 & 29.21 & 32.58 & 28.60 & 48.12 \\
Proposed \textbf{UNIQUE}& 1024 & \textbf{36.72} & \textbf{35.77} & \textbf{37.98} & \textbf{31.02} & \textbf{32.78} & \textbf{30.75} & \textbf{48.34} \\
\hline
Quest          & 512  & 21.72 & 21.09 & 21.84 & 18.87 & 16.18 & 17.61 & 29.31 \\
H2O            & 512  & 29.04 & 27.96 & 30.44 & 23.61 & 28.15 & 22.20 & 39.13 \\
InfLLM         & 512  & 34.99 & 33.82 & 36.21 & 28.96 & 32.38 & 27.31 & 46.74 \\
Proposed \textbf{UNIQUE}& 512  & \textbf{36.58} & \textbf{35.22} & \textbf{38.23} & \textbf{30.86} & \textbf{33.75} & \textbf{30.37} & \textbf{47.67} \\
\hline
Quest          & 256  & 15.41 & 15.65 & 14.67 & 14.78 & 11.34 & 12.66 & 19.29 \\
H2O            & 256  & 26.47 & 25.76 & 26.96 & 19.05 & 25.42 & 19.58 & 37.67 \\
InfLLM         & 256  & 33.51 & 32.87 & 34.08 & 27.59 & 30.39 & 25.62 & 45.40 \\
\textbf{UNIQUE}& 256  & \textbf{35.91} & \textbf{35.11} & \textbf{36.88} & \textbf{30.23} & \textbf{33.38} & \textbf{29.11} & \textbf{46.94} \\
\hline
Quest          & 128  & 8.36 & 9.56 & 7.75 & 9.85 & 5.59 & 6.73 & 10.55 \\
H2O            & 128  & 24.25 & 23.74 & 24.68 & 14.10 & 24.71 & 18.85 & 36.35 \\
InfLLM         & 128  & 30.88 & 30.39 & 31.11 & 25.50 & 29.89 & 23.38 & 40.49 \\
Propsoed \textbf{UNIQUE}& 128  & \textbf{34.70} & \textbf{33.47} & \textbf{36.41} & \textbf{28.90} & \textbf{31.97} & \textbf{29.00} & \textbf{45.80} \\
\Xhline{3\arrayrulewidth}
\end{tabular}
\vspace{-0.2cm}
\caption{Training-free LongBench-Pro results on
Ministral-3-8B-Instruct-2512 (higher is better, $\uparrow$).
\textbf{Budget} denotes the number of KV tokens each query is allowed
to attend to at decoding time; full attention attends to all KV tokens.}
\label{tab:longbench_trainfree}
\end{table*}

\begin{table*}[t]
\centering
\small
\renewcommand\arraystretch{1.1}
\setlength{\tabcolsep}{2.6pt}
\begin{tabular}{l|c|cccccccccccccc}
\Xhline{3\arrayrulewidth}
\textbf{Method} & \textbf{Budget}
& \textbf{SG1} & \textbf{SG2} & \textbf{SG3}
& \textbf{MK1} & \textbf{MK2} & \textbf{MK3}
& \textbf{MV} & \textbf{MQ}
& \textbf{VT} & \textbf{CWE} & \textbf{FWE}
& \textbf{QA1} & \textbf{QA2} & \textbf{AVG} \\
\hline
Full attention & --
& 100.00 & 99.60 & 99.60
& 99.40  & 100.00 & 99.80
& 98.55  & 99.05
& 99.96  & 88.46  & 93.27
& 75.80  & 57.40  & 93.15 \\
\hline
Quest                    & 2048
& \textbf{100.00} & 98.80 & 97.00
& 99.20  & 99.80  & 97.60
& \textbf{99.45} & \textbf{99.30}
& 99.88  & 77.80  & 90.67
& 75.00  & 57.20  & 91.67 \\
H2O                      & 2048
& \textbf{100.00} & \textbf{99.80} & 31.00
& \textbf{99.60} & 97.80  & 1.80
& 92.65  & 65.20
& 98.40  & 42.96  & 90.80
& 74.40  & 56.80  & 73.17 \\
InfLLM                   & 2048
& \textbf{100.00} & 94.00 & 85.20
& 95.20  & 99.80  & 96.40
& 93.00  & 88.45
& 94.08  & 22.10  & 83.80
& 74.60  & 57.20  & 83.37 \\
\textbf{UNIQUE} & 2048
& \textbf{100.00} & \textbf{99.80} & \textbf{99.20}
& \textbf{99.60} & \textbf{100.00} & \textbf{99.80}
& 99.00  & 99.15
& \textbf{99.96} & \textbf{82.28} & \textbf{91.07}
& \textbf{75.60} & \textbf{57.40} & \textbf{92.53} \\
\hline
Quest                    & 512
& 99.40 & 94.60 & 91.00
& 97.20 & 95.80 & 53.00
& 96.55 & 96.65
& 99.00 & 53.24 & 87.60
& 74.00 & 55.80 & 84.14 \\
H2O                      & 512
& 99.80 & 98.00 & 2.60
& 97.60 & 84.60 & 0.00
& 44.45 & 14.65
& 66.88 & 20.94 & 68.33
& \textbf{74.40} & 55.80 & 56.00 \\
InfLLM                   & 512
& 92.60 & 86.00 & 76.00
& 88.60 & 98.40 & 43.80
& 85.60 & 79.40
& 87.52 & 10.12 & 74.40
& 74.20 & 56.20 & 73.30 \\
\textbf{UNIQUE} & 512
& \textbf{100.00} & \textbf{99.80} & \textbf{99.20}
& \textbf{99.80} & \textbf{100.00} & \textbf{99.60}
& \textbf{99.55} & \textbf{99.30}
& \textbf{99.96} & \textbf{62.90} & \textbf{88.67}
& 73.60 & \textbf{57.80} & \textbf{90.78} \\
\Xhline{3\arrayrulewidth}
\end{tabular}
\vspace{-0.2cm}
\caption{Training-free RULER-32K results on
Ministral-3-8B-Instruct-2512 (higher is better, $\uparrow$).
\textbf{Budget} denotes the number of KV tokens each query is allowed
to attend to at decoding time; full attention attends to all KV tokens.}
\label{tab:ruler_trainfree}
\end{table*}

\subsection{Setup}
\label{sec:setup}

UNIQUE is evaluated on two modalities, long-context text understanding and long-form automatic speech recognition (ASR).
For text, training-free results are reported on LongBench-Pro~\citep{chen2026longbench}, a bilingual benchmark of 1{,}500 English and Chinese samples covering 11 primary and 25 secondary tasks with inputs up to 256K tokens, and on the 32K subset of RULER~\citep{hsieh2024ruler}, which spans retrieval, multi-hop tracing, aggregation, and question answering. Sparsity-aware fine-tuning uses LongAlign-10k~\citep{DBLP:conf/emnlp/BaiLZHQH0DL24}, a primarily English long-context corpus of about 10K instances; fine-tuned models are therefore evaluated on the English subset of LongBench-Pro. For speech, 69K hours of Portuguese ASR data (utterances up to one hour) are used to both train the dense speech LLM and conduct sparsity-aware fine-tuning, and evaluation is performed on long-form Portuguese ASR test sets of 10-minute utterances.

Ministral-3-8B-Instruct-2512~\citep{liu2026ministral} is used as the text LLM. The speech LLM uses Qwen3-8B~\citep{yang2025qwen3} as the language backbone, coupled with a Conformer speech encoder.

A series of popular sparse-attention methods of both flavors are implemented for comparison with the proposed UNIQUE. The \textbf{training-free} methods include H2O~\citep{DBLP:conf/nips/Zhang00CZC0TRBW23}, Quest~\citep{DBLP:conf/icml/TangZZXKH24}, and InfLLM~\citep{DBLP:conf/nips/XiaoZ0XLZ0024}; the \textbf{trainable} methods include InfLLM-v2~\citep{zhao2025infllm} and DeepSeek Sparse Attention (DSA)~\citep{liu2025deepseek}. All methods share the same LLM backbone and token budget as UNIQUE for a fair comparison, with the budget set to $512$ unless otherwise stated. More details refer to Appendix~\ref{app:hyper}.

\subsection{Long-Context Text Understanding}       
\paragraph{Training-free results}
Table~\ref{tab:longbench_trainfree} and Table~\ref{tab:ruler_trainfree}
report LongBench-Pro and RULER-32K accuracy of
Ministral-3-8B-Instruct-2512 under per-query KV budgets from $128$ to
$2048$ pages. The two benchmarks expose complementary failure modes
of prior methods. Quest~\citep{DBLP:conf/icml/TangZZXKH24}, which estimates page importance from
elementwise min/max of keys, remains competitive on RULER at all
budgets ($91.67$ at $2048$, $84.14$ at $512$) but degrades sharply on
LongBench-Pro ($21.72$ at $512$), indicating that min/max bounds
are adequate for needle-style retrieval yet miss the broader
reasoning context required by realistic long-context tasks. H2O~\citep{DBLP:conf/nips/Zhang00CZC0TRBW23} shows
the opposite pattern: its token-level eviction holds up on
LongBench-Pro ($29.04$ at $512$) but collapses on RULER
($56.00$ at $512$, with $0$ on \textsc{niah\_multikey\_3} (MK3)), because
permanently dropped KV cannot be recovered once a needle query
arrives. InfLLM~\citep{DBLP:conf/nips/XiaoZ0XLZ0024} is the strongest baseline on LongBench-Pro
($34.99$ at $512$) but trails Quest on RULER ($73.30$ vs.\ $84.14$).
UNIQUE is the only one that is uniformly strong on both: at the
default $512$-page budget it reaches $36.58$ on LongBench-Pro and
$90.78$ on RULER, recovering $97.0\%$ and $97.5\%$ of full attention
while outperforming the best baseline by $1.59$ and $6.64$ points
respectively. The advantage is consistent across languages and
difficulty levels and persists at the tighter $128$-page budget,
where competing methods lose $40\text{--}90\%$ of their full-attention
score.

\begin{table}[t]
\centering
\small
\renewcommand\arraystretch{1.1}
\setlength{\tabcolsep}{9.5pt}
\begin{tabular}{l|c|c}
\Xhline{3\arrayrulewidth}
\textbf{Method} & \textbf{Budget} & \textbf{English Split} ($\uparrow$) \\
\hline
Full attention                    & --   & 36.55 \\
\hline
InfLLM-v2                         & 512  & 36.11 \\
DSA                               & 512  & 36.79 \\
Proposed \textbf{UNIQUE}          & 512  & \textbf{37.25} \\
\Xhline{3\arrayrulewidth}
\end{tabular}
\vspace{-0.2cm}
\caption{Sparsity-aware fine-tuning results on the English split of
LongBench-Pro. All sparse methods and the full-attention baseline are
LoRA fine-tuned on LongAlign-10k from the same
Ministral-3-8B-Instruct-2512 checkpoint. \textbf{Budget} denotes the
number of KV tokens each query is allowed to attend to at decoding
time; full attention attends to all KV tokens.}
\label{tab:longbench_ft}
\end{table}

\paragraph{Sparsity-aware fine-tuning}
Table~\ref{tab:longbench_ft} reports LongBench-Pro English scores at
the default $512$-page budget after LoRA fine-tuning on
LongAlign-10k from the same Ministral-3-8B-Instruct-2512 checkpoint.
DSA~\citep{liu2025deepseek} was originally designed for MLA; this paper adapts it to GQA by
extending its lightning indexer from the MQA structure to a GQA
structure, so that each KV head has its own indexer and selects pages
independently within its group. All three sparse variants reach scores
close to the fine-tuned dense baseline ($36.55$); notably,
InfLLM-v2~\citep{zhao2025infllm} jumps to $36.11$ from its $33.82$ training-free counterpart
in Table~\ref{tab:longbench_trainfree}, largely closing the gap left by training-free top-$k$.
DSA ($36.79$) and UNIQUE ($37.25$) further surpass the
dense baseline, suggesting that sparsity-aware training even removes
some of the noise introduced by full-context attention. Among the
three, UNIQUE attains the highest accuracy while keeping the cleanest
training recipe: unlike DSA, UNIQUE requires neither an auxiliary
distillation loss against the dense attention distribution nor
materializing per-query attention scores from a separate indexer.

\subsection{Long-Form Speech Recognition}

\begin{table}[t]
\centering
\small
\renewcommand\arraystretch{1.1}
\setlength{\tabcolsep}{3.5pt}
\begin{tabular}{l|cc|cc|c}
\Xhline{3\arrayrulewidth}
& \multicolumn{2}{c|}{\textbf{Brazilian}}
& \multicolumn{2}{c|}{\textbf{European}}
& \\
\textbf{Method}
& \textbf{EER} & \textbf{WER}
& \textbf{EER} & \textbf{WER}
& \textbf{Avg} \\
\hline
Full attention
& 18.95 & 20.97 & 13.22 & 19.85 & 18.25 \\
\hline
Quest
& 22.97 & 23.02 & 19.37 & 22.30 & 21.92 \\
InfLLM
& 56.02 & 48.19 & 69.53 & 57.12 & 57.72 \\
Proposed \textbf{UNIQUE}
& \textbf{18.91} & \textbf{21.53} & \textbf{13.66} & \textbf{20.31} & \textbf{18.60} \\
\Xhline{3\arrayrulewidth}
\end{tabular}
\vspace{-0.2cm}
\caption{Training-free long-form ASR results (lower is better, $\downarrow$) on
Brazilian and European Portuguese long-form ($\sim$10-minute
utterances) test sets at KV budget $512$. Entity Error Rate (EER) requires every word of a reference entity to be
transcribed verbatim.}
\label{tab:asr_trainfree}
\end{table}

\paragraph{Training-free results}
Long-form ASR is a particularly demanding stress test for sparse
attention: a single dropped phrase or hallucinated word raises word error rate (WER),
and the model cannot paraphrase its way around a missing page.
Beyond standard WER, this paper also reports Entity Error Rate (EER,
defined as $1{-}$Entity Recall), which counts a reference entity as
correctly recognized only when every constituent word is transcribed
verbatim, providing a stricter measure of semantic integrity on
information-dense long-form speech.
Table~\ref{tab:asr_trainfree} reports results on Brazilian and
European Portuguese long-form ($\sim$10-minute) test sets at a
$512$-page KV budget. H2O fails to produce usable transcriptions
under this setting and is omitted from the table.
InfLLM degrades the full-attention baseline by $40$--$55$ absolute
points on both metrics, suggesting that its selection heuristics,
tuned for text, may need adaptation for other modalities. Quest is the
strongest baseline ($21.92$ in average), still
$3.67$ points behind full attention. UNIQUE is the only method that
matches dense attention on speech: its macro-average ($18.60$) is
within $0.35$ points of the dense baseline ($18.25$), and on
Brazilian EER it even slightly improves over dense ($18.91$ vs.\
$18.95$). Together with the text-domain results in
Tables~\ref{tab:longbench_trainfree} and~\ref{tab:ruler_trainfree},
this confirms that the mean-plus-std page score is modality-agnostic
and remains reliable on inputs whose statistics differ substantially
from natural text.

\begin{table}[t]
\centering
\small
\renewcommand\arraystretch{1.1}
\setlength{\tabcolsep}{3.5pt}
\begin{tabular}{l|cc|cc|c}
\Xhline{3\arrayrulewidth}
& \multicolumn{2}{c|}{\textbf{Brazilian}}
& \multicolumn{2}{c|}{\textbf{European}}
& \\
\textbf{Method}
& \textbf{EER} & \textbf{WER}
& \textbf{EER} & \textbf{WER}
& \textbf{Avg} \\
\hline
Full attention
& 18.95 & 20.97 & 13.22 & 19.85 & 18.25 \\
\hline
InfLLM-v2
& 20.07 & 22.25 & 13.92 & 20.15 & 19.10 \\
DSA
& 20.00 & 27.86 & 13.47 & 22.29 & 20.91 \\
Proposed \textbf{UNIQUE}
& \textbf{18.54} & \textbf{20.75} & \textbf{13.09} & \textbf{19.17} & \textbf{17.89} \\
\Xhline{3\arrayrulewidth}
\end{tabular}
\vspace{-0.2cm}
\caption{Sparsity-aware fine-tuning ASR results ($\downarrow$) on the
same test sets as Table~\ref{tab:asr_trainfree} at KV budget $512$.}
\vspace{-0.2cm}
\label{tab:asr_ft}
\end{table}

\paragraph{Sparsity-aware fine-tuning}

The speech LLM is itself fine-tuned from a text LLM, so sparsity-aware adaptation uses the same ASR data and recipe as the dense run; the full-attention numbers therefore match Table~\ref{tab:asr_trainfree} and the comparison among sparse methods is fair. Both InfLLM-v2 and DSA perform well after fine-tuning, but UNIQUE reaches $17.89$ macro-average, beating every baseline on every split and even slightly surpassing dense ($18.25$), without the local-window heuristic of InfLLM-v2 whose modality-specific prior may not always transfer. DSA closes most of the EER gap, showing that distilling the indexer recovers salient entity information, yet its WER remains noticeably higher; long-form ASR appears to expose the limits of attention distillation when even non-salient acoustic context must be retained for accurate transcription.

\subsection{Decoding Efficiency}
\label{sec:efficiency}

\begin{figure}[t]
\centering
\includegraphics[width=1.0\linewidth]{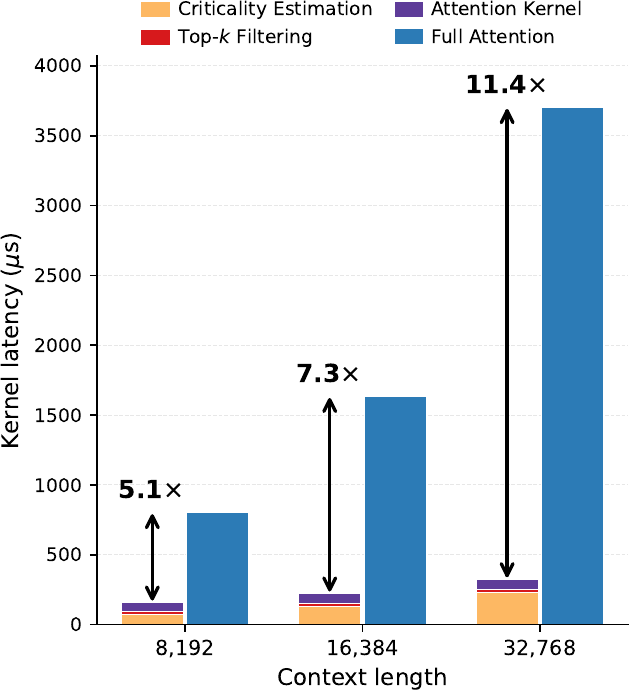}
\vspace{-0.71cm}
\caption{UNIQUE self-attention latency, which is broken
down into criticality estimation, top-$k$ filtering, and attention kernel; arrows indicate UNIQUE speedup over dense FlashInfer.}
\label{fig:breakdown}
\end{figure}

\paragraph{Attention kernel speedup}
Figure~\ref{fig:breakdown} reports the self-attention latency on a H100 with $32$ query heads, $8$ KV heads and head dimension
$128$, at page size $S{=}8$, KV budget $512$, and batch size $80$.
UNIQUE reaches a $11.4\times$ speedup over dense FlashInfer at $32$K
context, 
and the gap widens with context length: the fixed-budget sparse pipeline grows sublinearly while dense scales linearly.

\begin{figure}[t]
\centering
\includegraphics[width=0.93\linewidth]{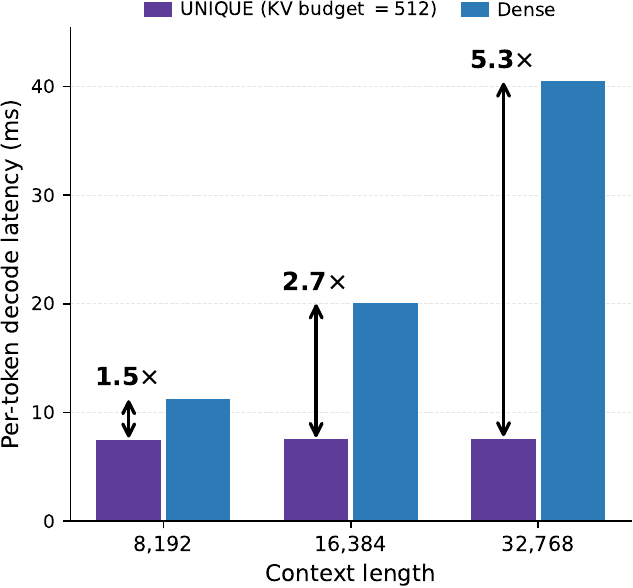}
\vspace{-0.2cm}
\caption{End-to-end per-token decoding latency under vLLM, batch
$80$, KV budget $512$. Arrows mark UNIQUE speedup over the dense
baseline.}
\label{fig:e2e}
\end{figure}

\paragraph{End-to-end speedup.}
Figure~\ref{fig:e2e} reports the end-to-end per-token decoding
latency of UNIQUE on a $16$-layer LLM with $32$ query
heads, $8$ KV heads and head dimension $64$, at batch size $80$
with the default $512$-token KV budget; both UNIQUE and the dense
baseline run on vLLM~\citep{DBLP:conf/sosp/KwonLZ0ZY0ZS23}. At $32$K
context UNIQUE reaches up to $5.3\times$ end-to-end speedup over
dense attention. 
The sparse pipeline keeps the per-token latency nearly constant
because attention occupies only a small share of decoding in this
model, masking the criticality estimator's linear growth in context.

\subsection{Ablations}
\label{sec:ablation}

\begin{table}[t]
\centering
\small
\renewcommand\arraystretch{1.1}
\setlength{\tabcolsep}{4.3pt}
\begin{tabular}{l|cc|cc|c}
\Xhline{3\arrayrulewidth}
& \multicolumn{2}{c|}{\textbf{Brazilian}}
& \multicolumn{2}{c|}{\textbf{European}}
& \\
\textbf{Scoring} & \textbf{EER} & \textbf{WER}
& \textbf{EER} & \textbf{WER} & \textbf{Avg} \\
\hline
UNIQUE w/o std   & \textbf{18.11} & 22.66 & 14.52 & 21.14 & 19.11 \\
UNIQUE & 18.91 & \textbf{21.53}
& \textbf{13.66} & \textbf{20.31} & \textbf{18.60} \\
\Xhline{3\arrayrulewidth}
\end{tabular}
\vspace{-0.2cm}
\caption{Training-free ablation of the std offset on the same ASR
test sets as Table~\ref{tab:asr_trainfree} (\%, $\downarrow$).}
\vspace{-0.15cm}
\label{tab:abl_offset}
\end{table}

\begin{table}[t]
\centering
\small
\renewcommand\arraystretch{1.1}
\setlength{\tabcolsep}{4.5pt}
\begin{tabular}{l|cc|cc|c}
\Xhline{3\arrayrulewidth}
& \multicolumn{2}{c|}{\textbf{Brazilian}}
& \multicolumn{2}{c|}{\textbf{European}}
& \\
\textbf{Mask} & \textbf{EER} & \textbf{WER}
& \textbf{EER} & \textbf{WER} & \textbf{Avg} \\
\hline
UNIQUE & \textbf{18.54} & \textbf{20.75}
& \textbf{13.09} & \textbf{19.17} & \textbf{17.89} \\
\, \, w/ hard mask & 18.89 & 20.84 & 13.83 & 19.73 & 18.32 \\

\Xhline{3\arrayrulewidth}
\end{tabular}
\vspace{-0.2cm}
\caption{Sparsity-aware fine-tuning ablation of the soft mask on the
same ASR test sets as Table~\ref{tab:asr_trainfree} (\%, $\downarrow$).}
\label{tab:abl_softmask}
\end{table}  

\paragraph{Effect of the std offset}
Table~\ref{tab:abl_offset} compares the training-free UNIQUE scoring (Eq.~\ref{eq:score}) against its mean-only ablation
$\bm{q} \cdot \bm{\mathrm{mean}}_p$ on long-form ASR with a $512$-token
KV budget. Removing the std term raises the macro-average from $18.60$
to $19.11$, with the largest regression on Brazilian WER ($21.53
\rightarrow 22.66$), confirming that the std offset captures useful
within-page variance that pure mean scoring under-weights.

\paragraph{Effect of the soft mask}
Table~\ref{tab:abl_softmask} ablates the sigmoid soft mask in
sparsity-aware fine-tuning against a hard top-$k$ mask under
otherwise identical settings. The soft mask reduces the macro-average
from $18.32$ to $17.89$ and improves every individual split.
The gain comes from differentiability: the soft mask propagates
gradients into the criticality estimator so the page-selection
process itself adapts, whereas the hard mask only lets the model
adapt to the already-selected pages.

\section{Conclusions}
\label{sec:conclusion}
This paper proposes UNIQUE, a universal top-$k$ sparse attention
that supports both training-free inference and sparsity-aware
fine-tuning. UNIQUE scores KV pages by a simple yet effective
mean-plus-std formula, and adapts the model via a soft-mask
training scheme without auxiliary losses or model changes.
On LongBench-Pro, RULER-32K and long-form ASR,
UNIQUE is the only training-free method that matches dense
attention across retrieval-style and realistic long-context tasks
in text and speech; sparsity-aware fine-tuning further closes
the residual gap. On a single H100, UNIQUE reaches $11.4\times$
attention-kernel and $5.3\times$ end-to-end speedup at $32$K context, validating a single, modality-agnostic sparse attention design.


\newpage
\section*{Limitations}

KV-cache reduction is a broad and fast-moving area. This paper
compares UNIQUE against representative training-free baselines
(Quest, InfLLM, H2O) and trainable baselines (InfLLM-v2, DSA), but
many other directions, including KV quantization and low-rank
compression, are orthogonal and not compared here. A more
exhaustive empirical sweep is left to future work.

This paper evaluates UNIQUE on text and speech LLMs, two of the
modalities most affected by long-context inference. Extending the
study to vision and vision-language models, which lie outside our
expertise, is left to future work.

This paper considers UNIQUE under the fine-tuning regime, which
is the setting most directly relevant to deployment and which
allows comparison with prior fine-tunable baselines such as
InfLLM-v2 and DSA. Methods that are designed for pre-training
from scratch, such as Native Sparse Attention,
are therefore not included in the comparison: they cannot be
fairly evaluated under fine-tuning alone, and a from-scratch LLM
pre-training is beyond our available compute. Although the
design of UNIQUE itself does not preclude such a pre-training
setting, that scenario is left to future work.

\section*{Ethical Considerations}

This paper proposes UNIQUE to lower the inference cost of
long-context LLMs in both text and speech. Today such inference
remains expensive and is therefore most accessible to
well-resourced groups; methods that reduce the per-token cost,
such as UNIQUE, can broaden access to long-context capabilities
and reduce the energy footprint of large-scale deployment.
UNIQUE does not introduce additional sources of bias beyond those
already present in the pre-trained LLM checkpoints and training
data on which it is applied. All datasets, pre-trained models and software libraries used in this work are accessed under their stated terms for academic
research, and the evaluation data does not contain personally
identifying information or offensive content. AI assistants were used only to polish the writing of this paper.



\bibliography{custom}

\appendix

\begin{table*}[t!]
\centering
\begin{tabular}{c c c c c}
\toprule
\textbf{KV len $N$} & \textbf{Pages $P$} & \textbf{Naive Implementation} & \textbf{Fused kernel} & \textbf{Speedup} \\
\midrule
8K   & 1{,}024  & 0.0670 ms & {0.0388 ms} & {1.73$\times$} \\
16K  & 2{,}048  & 0.0994 ms & {0.0589 ms} & {1.69$\times$} \\
32K  & 4{,}096  & 0.1597 ms & {0.0990 ms} & {1.61$\times$} \\
64K  & 8{,}192  & 0.2706 ms & {0.1804 ms} & {1.50$\times$} \\
128K & 16{,}384 & 0.4984 ms & {0.3422 ms} & {1.46$\times$} \\
\bottomrule
\end{tabular}
\caption{Per-step latency of the criticality estimation kernel on an NVIDIA H100 GPU. The fused kernel computes Eq.~\ref{eq:score_gqa} in a single CUDA launch; the naive implementation runs the same computation as three separate kernels.}
\label{tab:fused_kernel_bench}
\end{table*}

\begin{table*}[th!]
\centering
\begin{tabular}{c c c c c c}
\toprule
\textbf{Pages $P$} & \textbf{torch.topk} & \textbf{flashinfer.top\_k} & \textbf{Top-$k$ of UNIQUE} & \textbf{vs FlashInfer} & \textbf{vs PyTorch} \\
\midrule
1{,}024  & 0.031 ms & 0.026 ms & {0.013 ms} & {2.0$\times$} & {2.4$\times$} \\
2{,}048  & 0.038 ms & 0.026 ms & {0.013 ms} & {2.0$\times$} & {2.9$\times$} \\
4{,}096  & 0.068 ms & 0.025 ms & {0.014 ms} & {1.8$\times$} & {4.8$\times$} \\
8{,}192  & 0.076 ms & 0.027 ms & {0.018 ms} & {1.5$\times$} & {4.3$\times$} \\
16{,}384 & 0.102 ms & 0.032 ms & {0.022 ms} & {1.4$\times$} & {4.6$\times$} \\
\bottomrule
\end{tabular}
\caption{Per-step latency of the top-$k$ selection on an NVIDIA H100 GPU, with $k = 64$.}
\label{tab:radix_topk_bench}
\end{table*}

\section{Fused Criticality Estimation Kernel Benchmark}
\label{app:fused_kernel_bench}

This appendix reports a microbenchmark of the fused criticality estimation kernel of UNIQUE. The kernel implements Eq.~\ref{eq:score_gqa}: a batched matrix multiplication for the per-head raw scores, followed by an on-the-fly addition of the per-head outer-product offset $\lambda \, \|\bm{q}_g\|_2 \, \mathrm{std}_p$ and a max reduction across the $G$ query heads in a group. As a baseline, the same computation is performed naively as a batched matmul, a broadcast add, and a max reduction launched as three separate operations, which materializes the full $G \times P$ intermediate score tensor in HBM.

All measurements are taken on a single NVIDIA H100 GPU in bf16. The configuration matches the main experiments, with $H = 32$ query heads, $H_{kv} = 8$ KV heads (so $G = H/H_{kv} = 4$), head dimension $D = 128$, page size $S = 8$, $\lambda = 0.5$, and a batch of 32 decoding queries. The KV length $N$ is swept from 8K to 128K tokens, which gives a number of pages $P$ ranging from 1{,}024 to 16{,}384. Each entry is the median over 100 timed iterations after 20 warm-up runs.

Table~\ref{tab:fused_kernel_bench} reports the per-step latency of the criticality estimation step alone. The fused kernel is consistently faster than the naive baseline across all KV lengths, with up to 1.73$\times$ speedup at $N = 8$K. As $N$ grows, the speedup ratio decreases because the matrix multiplication itself becomes the dominant cost and leaves less room for the bias and reduction steps to amortize, but the absolute time saving keeps growing and reaches 0.156\,ms at $N = 128$K.

\section{Radix-based Top-$k$ Kernel Benchmark}
\label{app:radix_topk_bench}

This appendix reports a microbenchmark of the radix-based top-$k$ kernel of UNIQUE. The kernel takes a batch of bf16 score vectors of length $P$ and returns, for each vector, the indices of the $k$ largest entries via the two-round 8-bit radix selection described in Section~\ref{sec:inference}. As baselines, we compare against the top-$k$ routines in PyTorch (\texttt{torch.topk}) and FlashInfer (\texttt{flashinfer.top\_k}).

All measurements are taken on a single NVIDIA H100 GPU in bf16, with a batch of 32 decoding queries and $H_{kv} = 8$ KV heads, so each call performs $32 \times 8 = 256$ independent top-$k$ selections in parallel. The selection budget is fixed to $k = 64$, corresponding to a token budget of 512 under page size $S = 8$. The number of pages $P$ is swept from 1{,}024 to 16{,}384, covering KV lengths from 8K to 128K tokens. Each entry is the median over 300 timed iterations after 50 warm-up runs.

Table~\ref{tab:radix_topk_bench} reports the per-step latency of the top-$k$ selection alone. The radix-based kernel is consistently faster than both baselines across all $P$, with up to $2.0\times$ speedup over FlashInfer and up to $4.8\times$ speedup over \texttt{torch.topk}. The gap over \texttt{torch.topk} widens with $P$, reflecting the close-to-linear scaling of the radix kernel.

\section{Hyper-parameters}
\label{app:hyper}

Across all experiments, UNIQUE uses a page size of $S{=}8$. For the
DSA baseline implemented in this paper, the lightning indexer follows
the same GQA layout as the base LLM but with the head dimension
reduced to $1/8$ of the original (e.g.\ from $128$ to $16$), so that
its criticality-estimation cost is comparable to that of UNIQUE.

For text sparsity-aware fine-tuning on LongAlign-10k, Ministral-3-8B-Instruct-2512
is fine-tuned with LoRA (rank $r=64$, $\alpha=128$) on all attention
and MLP linear layers. Training runs for $5000$ steps with a single
sequence of length $65\,536$ per device, optimized by AdamW
($\beta_1{=}0.9$, $\beta_2{=}0.95$, $\epsilon{=}10^{-7}$, weight
decay $0.01$). The learning rate is $5\!\times\!10^{-6}$ with $100$
warm-up steps followed by cosine decay, gradient clipping at $1.0$,
and bf16 mixed precision with gradient checkpointing.

For speech sparsity-aware fine-tuning, a Qwen3-8B based speech LLM
is built by attaching a $24$-block Conformer audio encoder
(attention dim $1024$, $16$ heads, FFN $1536$, kernel size $3$, $8\times$ time reduction, T5 relative position bias) to the
language model. Fine-tuning runs on long-form Portuguese ASR for
$10{,}000$ steps, with sequences packed into $16{,}384$-token batches
and a global batch of $32$ via gradient accumulation; utterances
longer than $30$ minutes are filtered out due to GPU memory limit. LoRA (rank $r=320$, $\alpha=640$) is applied to all attention and MLP
linear layers. The learning rate is $4\!\times\!10^{-5}$ with $100$
warm-up steps and linear decay, and AdamW uses the same
hyperparameters as above. Training is run with gradient clipping at
$1.0$, DeepSpeed ZeRO-$1$, and bf16 mixed precision.

\end{document}